\documentclass{scspaperproc}

\usepackage{latexsym}
\usepackage{graphicx}
\usepackage{mathptmx}

\usepackage{amsmath}
\usepackage{amsfonts}
\usepackage{amssymb}
\usepackage{amsbsy}
\usepackage{amsthm}

\usepackage[pdftex,colorlinks=true,urlcolor=blue,citecolor=black,anchorcolor=black,linkcolor=black,bookmarks=false]{hyperref}

\usepackage{hyphenat}
\newtheoremstyle{scsthe}
{8pt}
{8pt}
{\it}
{}
{\bf}
{.}
{.5em}
{}

\theoremstyle{scsthe}

\begin{document}

%
%
\SCSpagesetup{Cheng and Ji}

\def\SCSconferencename{Annual Simulation Conference}

\def\SCSconferenceacro{ANNSIM'24}

\def\SCSpublicationyear{2024}


\def\SCSconferencedates{May 20-23}

\def\SCSconferencevenue{Perkins\&Will, Seattle, USA, Carnegie Mellon University, Pittsburgh, USA}

\title{Evaluating Panoramic 3D Estimation in Indoor Lighting Analysis}

\author[\authorrefmark{1}]{Zining Cheng}
\author[\authorrefmark{2}]{Guanzhou Ji}

\affil[\authorrefmark{1}]{Perkins\&Will, Seattle, USA}
\affil[ ]{\textit {zining.cheng@perkinswill.com}}

\affil[\authorrefmark{2}]{Carnegie Mellon University, Pittsburgh, USA}
\affil[ ]{\textit{gji@andrew.cmu.edu}}

\maketitle

\section*{Abstract}
This paper presents the use of panoramic 3D estimation in lighting simulation. Conventional lighting simulation necessitates detailed modeling as input, resulting in significant labor effort and time cost. The 3D layout estimation method directly takes a single panorama as input and generates a lighting simulation model with room geometry and window aperture. We evaluate the simulation results by comparing the luminance errors between on-site High Dynamic Range (HDR) photographs, 3D estimation model, and detailed model in panoramic representation and fisheye perspective. Given the selected scene, the results demonstrate the estimated room layout is reliable for lighting simulation. 

\textbf{Keywords:} HDR Photography, 3D Estimation, Panorama, Lighting Simulation.

\section{Introduction}
\label{sec:intro}

Accurate light estimation plays an important role in building design. High Dynamic Range (HDR) photography and computational lighting simulation are the two primary methods for indoor lighting analysis. The HDR photography method can capture a wide range of pixels from real-world scenes. However, on-site data collection requires manual input for capturing HDR photographs. Each captured image reflects the real-time scene radiance at a particular point of time. Also, in the conventional HDR workflow, the captured 2D image cannot be directly converted into 3D room geometry, thus limiting its application in 3D building analysis. Meanwhile, the standard modeling approach requires manual input from design professionals for individual building cases. 

3D estimation has been commonly used for understanding scene geometry from images. By using panoramic images as input, 3D estimation directly infers the complete room geometry. This process is fully generalizable without manual inputs, significantly minimizing labor and time costs in schematic design. However, when the estimated 3D model is used for indoor lighting simulation, room geometries are simplified to planar surfaces. The main hypothesis of this study is that the estimated 3D model can be used for indoor lighting simulation. The simulated results and error analysis will be discussed in this paper.

As shown in Table~\ref{table1:3_method}, this paper aims to evaluate panoramic 3D estimation as a method for generating 3D geometry from the real world and performing lighting analysis. Specifically, this paper makes the following technical contributions:
\begin{itemize}
\setlength\itemsep{0em}
\item We use a scalable approach to estimate panoramic 3D room Layout and construct a lighting simulation model.
\item We conclude the luminance errors between the HDR photograph, 3D estimation model, and detailed model from panoramic and fisheye perspectives. 
\item We analyze the discrepancies in glare metrics between 3D estimation and detailed modeling. 
\end{itemize}

\begin{table*}
    \caption{Comparision of Three Approaches for Indoor Lighting Analysis.}
    \centering
        \begin{tabular}{llll} \hline
        \bf   & \bf HDR Photography & \bf 3D Estimation & \bf Detailed Model\\ \hline
        Manual Input & Yes & No  & Yes \\ 
        Real-World Scene & Yes & No & No \\ 
        Room Geometry & No & Planar Surfaces & Detailed Geometry \\ 
        Light Customization & No & Yes & Yes \\ \hline
        \end{tabular}
    \label{table1:3_method}
\end{table*}

\section{Related Work}
\subsection{Climate-Based Lighting Simulation}
Physics-based lighting simulation utilizes building geometry, material definitions, and local weather data to create a virtual lighting environment. Indoor lighting distribution varies depending on geographic locations, sky conditions, and building orientations. Climate-Based Daylight Modeling (CBDM) facilitates computational lighting simulation under local weather conditions~\cite{mardaljevic2006examples}, by specifying time, geographic location~\cite{EnergyPlus2023weather}, and sky condition~\cite{standard2003spitial}. Daylight performance can be quantified using sustainable metrics~\cite{reinhart2006dynamic}. Computational lighting simulation flexibly renders hemispherical fisheye images at various indoor locations~\cite{ji2022view}, balancing daylight availability and occupant visual comfort~\cite{ji2020daylight,ji2022using}. Despite the time-consuming nature of traditional ray tracing engines~\cite{ward1994radiance} in lighting simulation, the conventional simulation process is expedited through parallel computing technologies~\cite{jones2017experimental}. Nevertheless, lighting simulation still requires manual inputs for detailed 3D room geometry. The geometric modeling process, typically conducted within the user interface, lacks generalizability across different scenes.

\subsection{3D Layout Estimation}
Recent studies focus on extracting semantic space information and floor representations~\cite{yang2022automated}, as well as recognizing detailed elements within floor plan layouts~\cite{zeng2019deep}, from 2D floor plans. Estimating the 3D room layout from a single image is also a common task in real-world scene understanding. One approach, based on the Manhattan world assumption~\cite{coughlan1999manhattan}, converts indoor panoramas into cubic maps and simplifies rooms as six-sided cubes~\cite{cheng2018cube}. Data-driven methods explore estimating a 360° room layout with individual planar surfaces using a single panorama as input~\cite{wang2021led2,zou2018layoutnet,zhang2014panocontext,yang2019dula}. Additionally, researchers investigate reconstructing detailed scene and furniture geometry from general 2D images~\cite{huang2018holistic,Izadinia2017,nie2020total3dunderstanding}.

\subsection{HDR Photography}
The range of scene radiance in the real-world scene spans from $10^{-3}$ $cd/m^{2}$ (starlight) to $10^{5}$ $cd/m^{2}$ (sunlight)~\cite{reinhard2010high}. The HDR technique captures a scene through multiple exposures and merges multiple images into one single HDR image. Given the captured HDR image reflects the relative luminance values of the real world, it is required to measure the absolute luminance value for each scene to recover the absolute scene luminance value~\cite{Debevec2008}. The photometric calibration is a means of radiometric self-calibration~\cite{mitsunaga1999radiometric}. The actual luminance ensures HDR images accurately represent real-time spatially-varying lighting conditions. However, capturing HDR images across various scenes and at different times is a time-consuming process.

\section{Methodology}
\begin{figure*}[!ht]
    \centering
    \includegraphics[width= .6\linewidth]{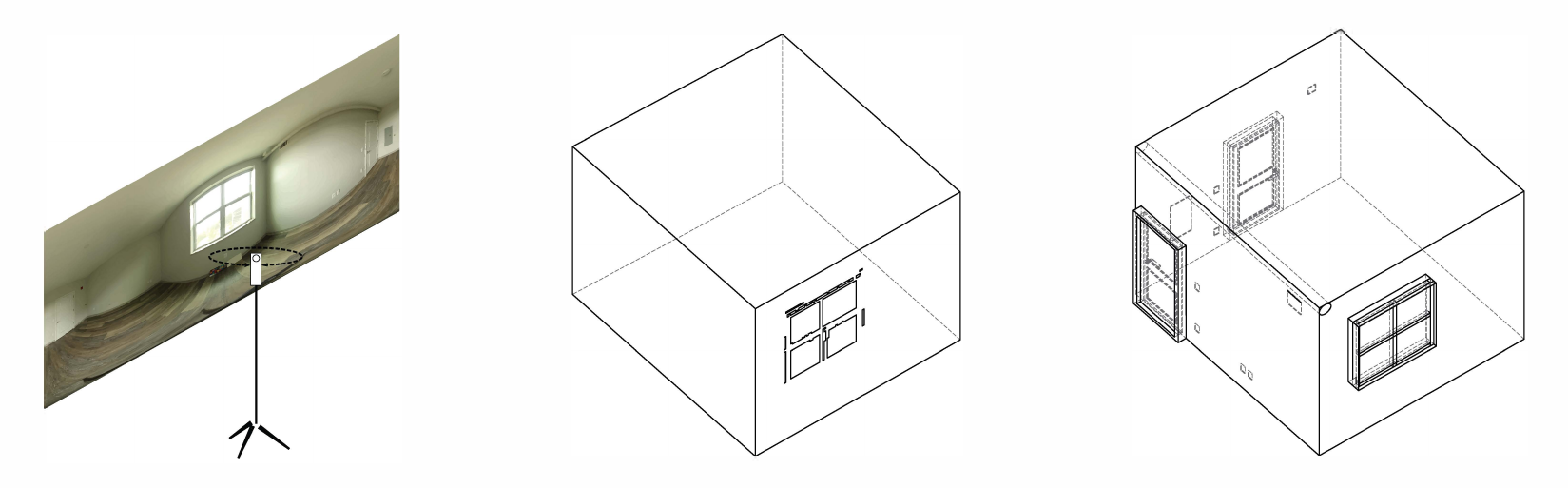}
    \caption{(left) HDR Photograph, (middle) 3D Estimation Model, and (right) Detailed Model.}
    \label{fig:diagram3method}
\end{figure*}

As shown in Figure.~\ref{fig:diagram3method}, three approaches for indoor lighting analysis will be compared in this study. This section presents the details of each approach and simulation settings for generating data. 

\subsection{Panoramic HDR Photography}
Compared to conventional fisheye images, a single panorama captures the complete content of the space from 360$^{\circ}$ directions. In this study, we focus on a single room and select panoramic HDR photographs at various time points and under different sky conditions from the calibrated panorama dataset~\cite{ji2023calih}. We will compare these calibrated HDR images with panoramic luminance maps generated from computational lighting simulations.

\begin{figure}
{
\centering
\includegraphics[width=0.6\textwidth]{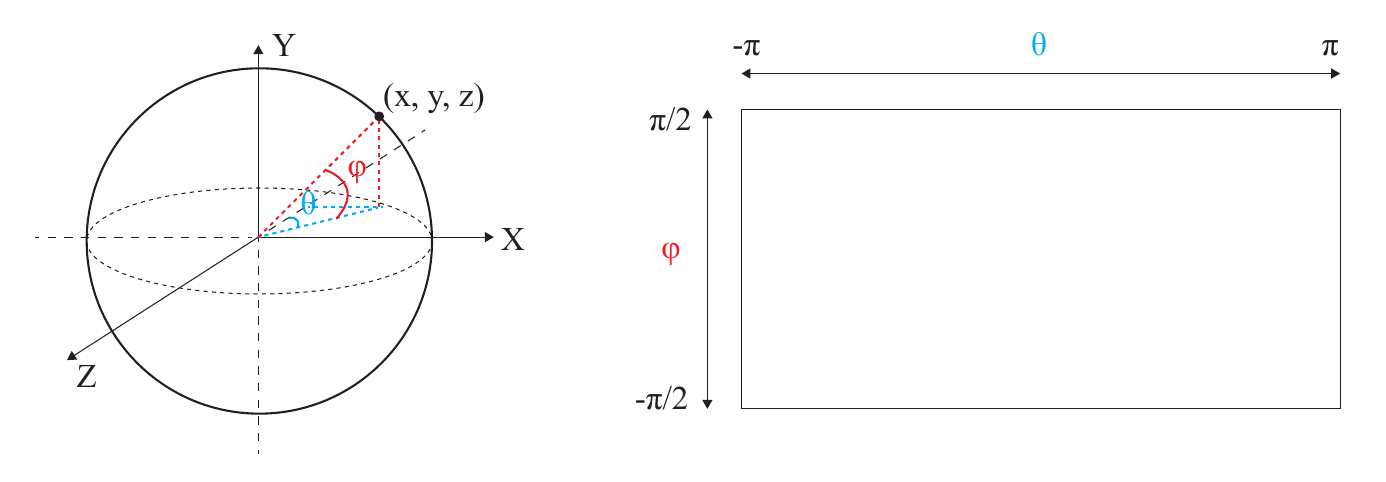}
\caption{(left) Spherical Coordinate and (right) Equirectanglar Representation.}
\label{fig:equ_coordinate}
}
\end{figure}

Panoramic photography captures a 3D scene and represents it in an equirectangular representation. In the spherical coordinate system (Figure.~\ref{fig:equ_coordinate} (left)), a single point on a unit sphere can be expressed by $\theta$ and $\phi$ \eqref{equation1}, where $\theta$ represents the Azimuth angle and $\phi$ indicates the Altitude angle. The range of $\theta$ is ($-\pi$, $+\pi$), while the range of $\phi$ is ($-0.5\pi$, $+ 0.5\pi$).

Meanwhile, given the known $x$, $y$, and $z$ values in the spherical coordinate, the $\theta'$ and $\phi'$ in equirectangular representation can also be obtained through \eqref{equation2}. Figure.~\ref{fig:equ_coordinate} (right) shows the mapping process when the 3D point set on the spherical coordinate system is projected onto the equirectangular representation. The transformation between a 180$^{\circ}$ fisheye image and the equirectangular image is invertible. Using the captured indoor panorama, we can modify the azimuth angle to produce fisheye images in different horizontal directions.

\begin{equation}\label{equation1}
    \begin{split}
        S(\theta, \phi) &= (x, y, z) ; \\
        x &= \cos(\phi) \sin(\theta) ; \\
        y &= \sin(\phi) ; \\
        z &= \cos(\phi) \cos(\theta).
    \end{split}
\end{equation}

\begin{equation}\label{equation2}
    \begin{split}
        \theta' &= arctan\frac{x}{z} ; \\
        \phi' &= arcsin\frac{y}{\sqrt{x^{2}+y^{2}+z^{2}}}. \\
    \end{split}
\end{equation}

\subsection{3D Estimation}
\begin{figure*}[!ht]
    \centering
    \includegraphics[width= 0.8\linewidth]{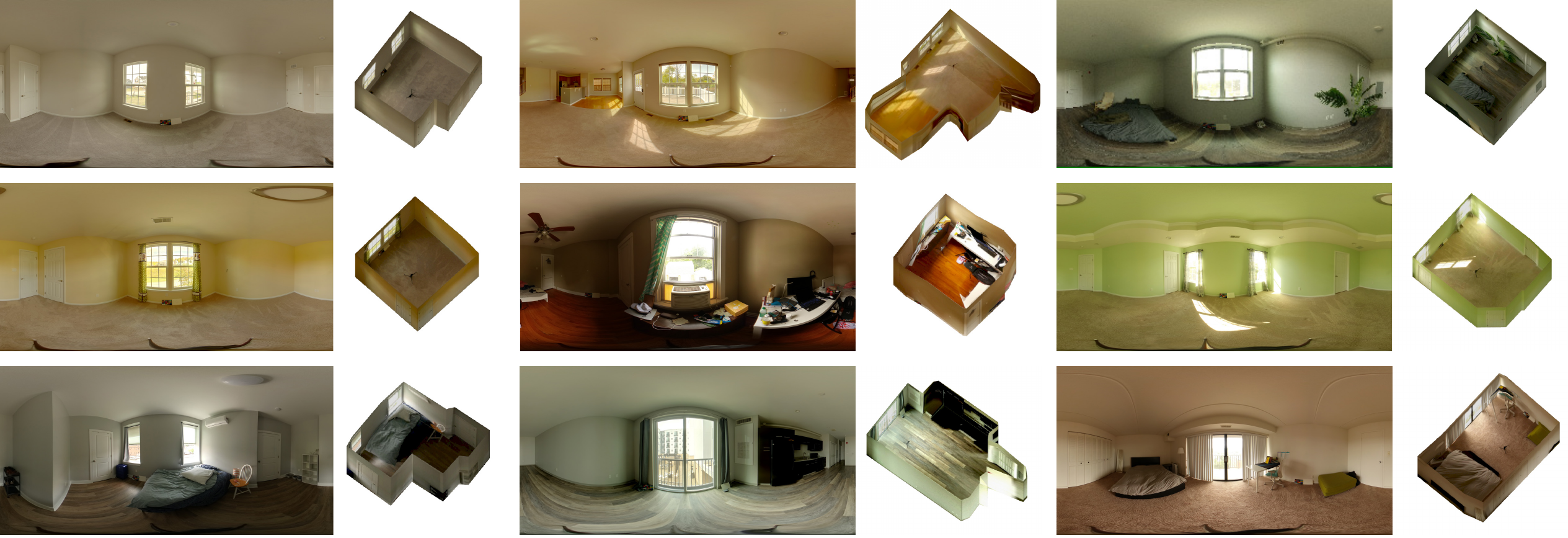}
    \caption{Examples of the estimated 3D Layouts from different indoor panoramas.}
    \label{fig:3D_reconstruct}
\end{figure*}

To generate 3D room geometry, the captured panorama will be used as an input to estimate indoor wall corners using the LED2Net model~\cite{wang2021led2}. Then, we use the estimated wall corners on the image to segment the panorama and construct the 3D room layout with individual planar surfaces (Figure.~\ref{fig:3D_reconstruct}). Each window, with its unique pattern and geometry, impacts the admission of light from the outdoor to the indoor space. To outline a detailed window pattern, we incorporate the window detection workflow~\cite{Ji2023virtual}. The modeling approach uses semantic segmentation~\cite{zhou2018semantic} to identify the window region and line detection~\cite{zhou2019end} to detect the window frames. The window aperture and estimated 3D geometry will be utilized for computational lighting simulation.

\subsection{Detailed Model}
\begin{figure}
    \centering
    \includegraphics[width=.60\linewidth]{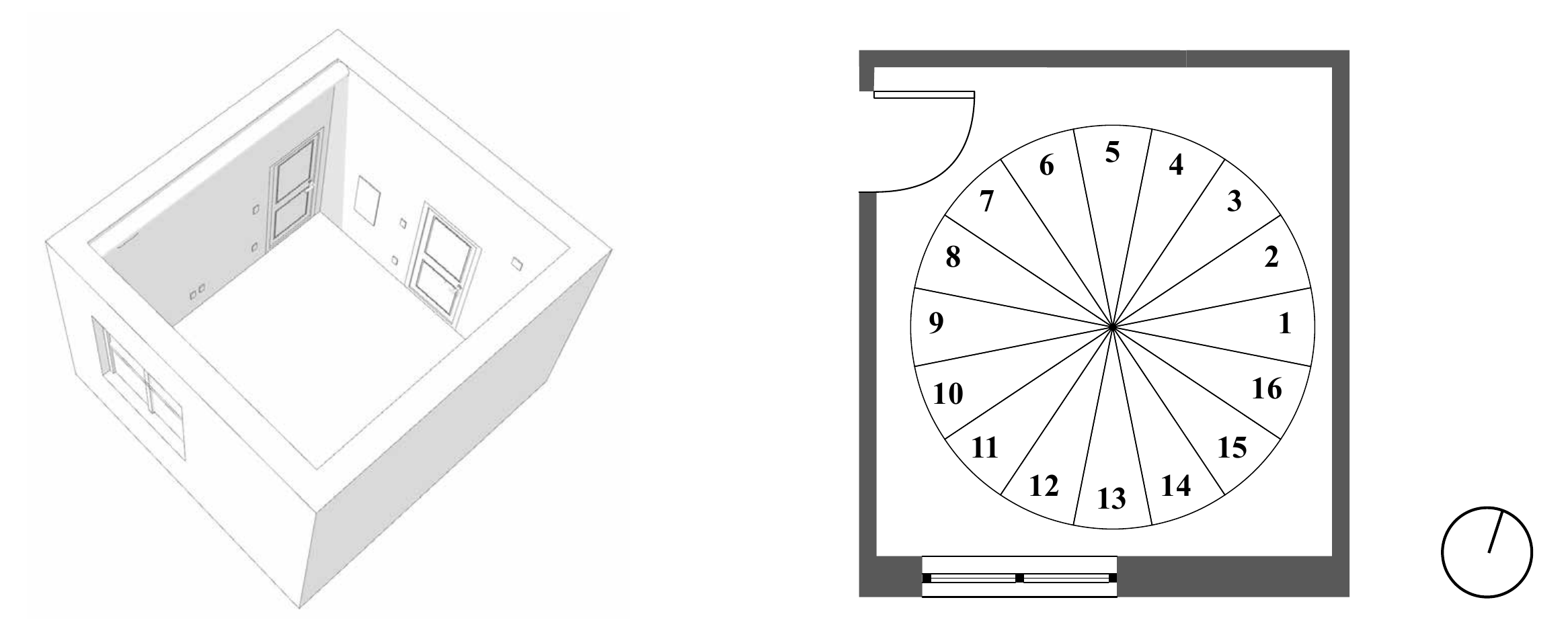}
    \caption{Detailed 3D geometry of the selected room and view divisions for glare analysis.}
    \label{fig:room}
\end{figure}

The selected space is an existing residential unit (Figure.~\ref{fig:room}), and the window faces 167$^{\circ}$ South orientation. The detailed scene model was built in Rhino~\cite{mcneel2020rhino} and Grasshopper plugin~\cite{gh_pliugin}. 
Compared to the estimated room geometry with planar surfaces, the detailed model, created through manual inputs, incorporates intricate indoor elements such as door frames, window frames, electrical outlets, smoke detectors, and plumbing, along with wall thickness.

\subsection{Simulation Settings}
The selected room is located in Pittsburgh, PA (latitude: 40.4406$^{\circ}$ N, longitude: 79.9959$^{\circ}$ W). The computational lighting simulation is performed in the Radiance-based~\cite{ward1994radiance} platform~\cite{CStudio}, using the local weather file (USA$^{\_}$PA$^{\_}$Pittsburgh.Intl.AP.725200$^{\_}$TMY3.epw) and Radiance parameters (ab: 8, lw: 0.01, samples per pixel: 100). We compare HDR photographs, 3D estimation, and detailed model under two different sky conditions (clear sky and overcast sky). We select time series HDR photographs from July 6th (clear sky) and June 22nd (overcast sky) from the calibrated panorama dataset~\cite{ji2023calih}. 

Daylight Glare Probability (DGP) is a glare metric that quantifies the percentage of occupants who perceive a scene as glary and quantifies the discomfort glare caused by natural sunlight. Higher DGP values correspond to a greater possibility for glare issues among building occupants~\cite{wienold2006evaluation}. The DGP classifies the result into four glare levels: imperceptible (\textgreater{} 0.35),  perceptible (0.35 - 0.40), disturbing (0.40 - 0.45), and intolerable (\textgreater{} 0.45) glare. For DGP analysis, the entire 360$^{\circ}$ view is segmented into sixteen divisions with 22.5$^{\circ}$ increments (Figure.~\ref{fig:room}). In the computational simulation for the 3D estimation model and detailed model, the hourly DGP results (from 8:30 to 16:30) from the target dates (March 21st, June 21st, September 21st, and December 21st) are generated from the complete annual glare simulation.

\section{Results}
\subsection{Panoramic Luminance Map}
The luminance maps from HDR photography, the 3D estimation model, and the detailed model will be compared side by side. Given $R$, $G$, and $B$ values in the captured indoor HDR image, each image is linearly calibrated by $k$ obtained from on-site luminance measurement, and the scene luminance is processed by per-pixel luminance calculation~\cite{inanici2006evaluation} \eqref{equation3}.

\begin{equation}\label{equation3}
  L =  k \cdot (0.2127 \cdot R + 0.7151 \cdot G + 0.0722 \cdot B) (cd/m^{2}) .
\end{equation}

\begin{figure*}[!ht]
    \centering
    \includegraphics[width=0.70\linewidth]{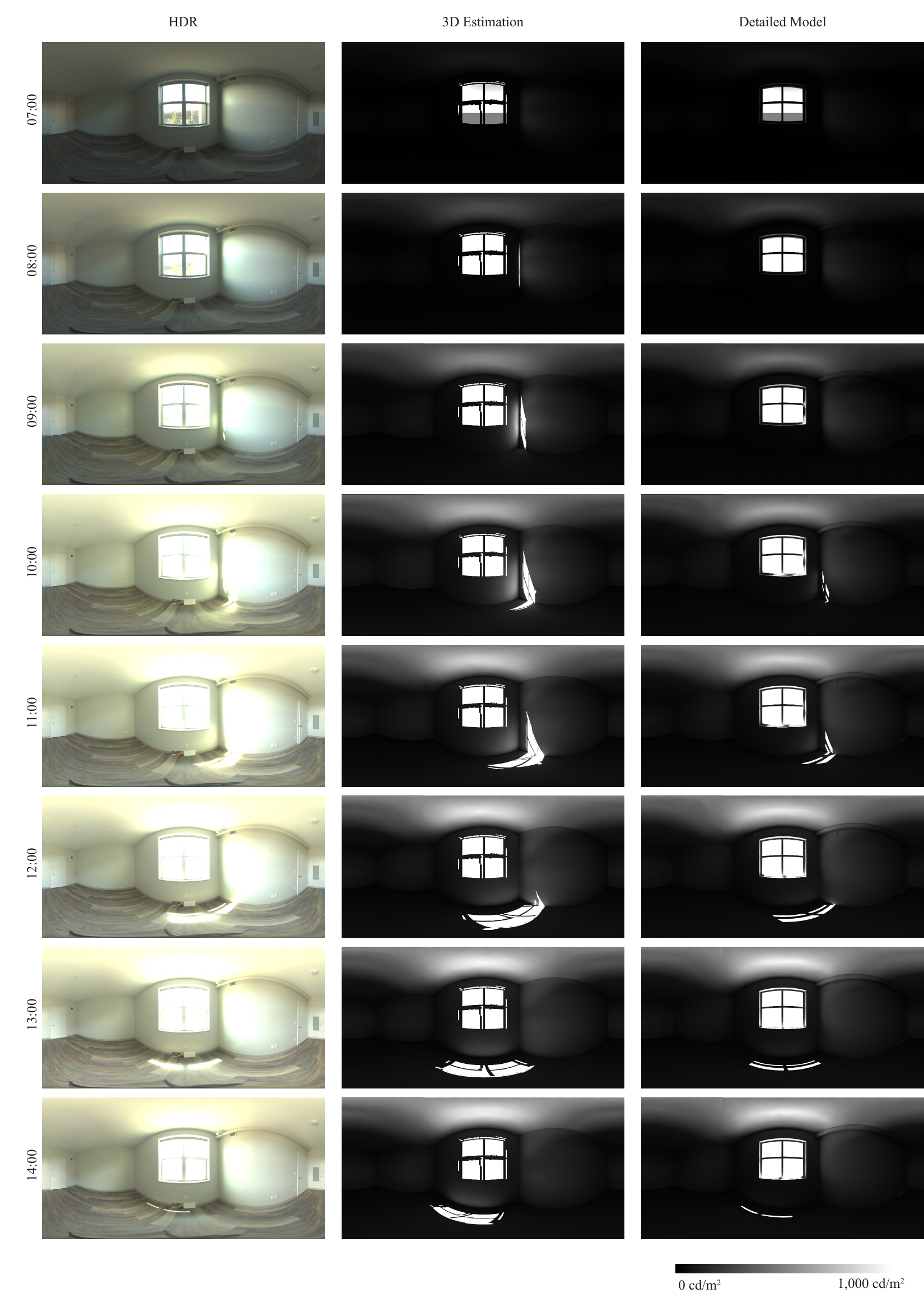}
    \caption{Panoramic HDR photographs displayed as Low Dynamic Range (LDR) images and luminance maps generated from 3D estimation and detailed modeling on July 6th under clear sky conditions.}
    \label{fig:result_pano_hdr}
\end{figure*}

As shown in Figure.~\ref{fig:result_pano_hdr}, we selected the HDR photographs on July 6th under clear sky conditions, with hourly intervals ranging from 07:00 to 14:00. This selection was made considering that direct sunlight cannot be observed from the scene after 14:00. Using the same time intervals, we obtained panoramic luminance maps from computational lighting simulation using the 3D estimation method and the detailed modeling approach, respectively. Compared to the detailed model method, the 3D estimation method brings larger regions of direct sunlight into the scene.

The HDR images captured the real-world textures and color characteristics, while the computational lighting simulation simplified the scene materials with uniform reflectance values. The 3D estimation method provides accurate room boundaries with individual planar surfaces. However, the detailed model incorporates wall thickness and indoor details (such as windows, sockets, smoke detectors, and water tubes).

\subsection{Panoramic Error Analysis}

\begin{figure*}[!ht]
    \centering
    \includegraphics[width=0.70\linewidth]{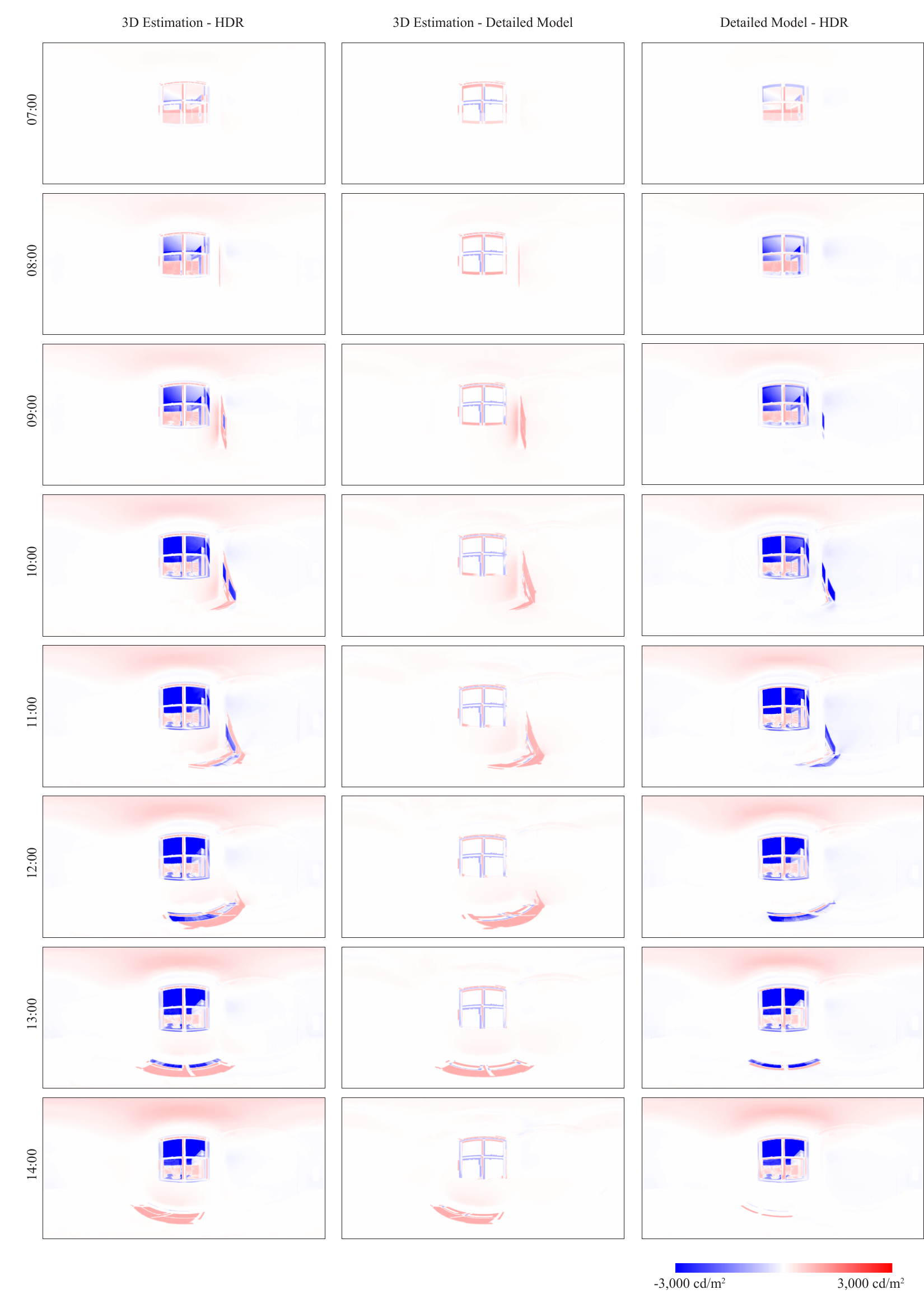}
    \caption{Luminance error between HDR photographs, 3D Estimation, and Detailed Model on July 6th under clear sky conditions.}
    \label{fig:error_0706}
\end{figure*}

\begin{figure*}[!ht]
    \centering
    \includegraphics[width=0.70\linewidth]{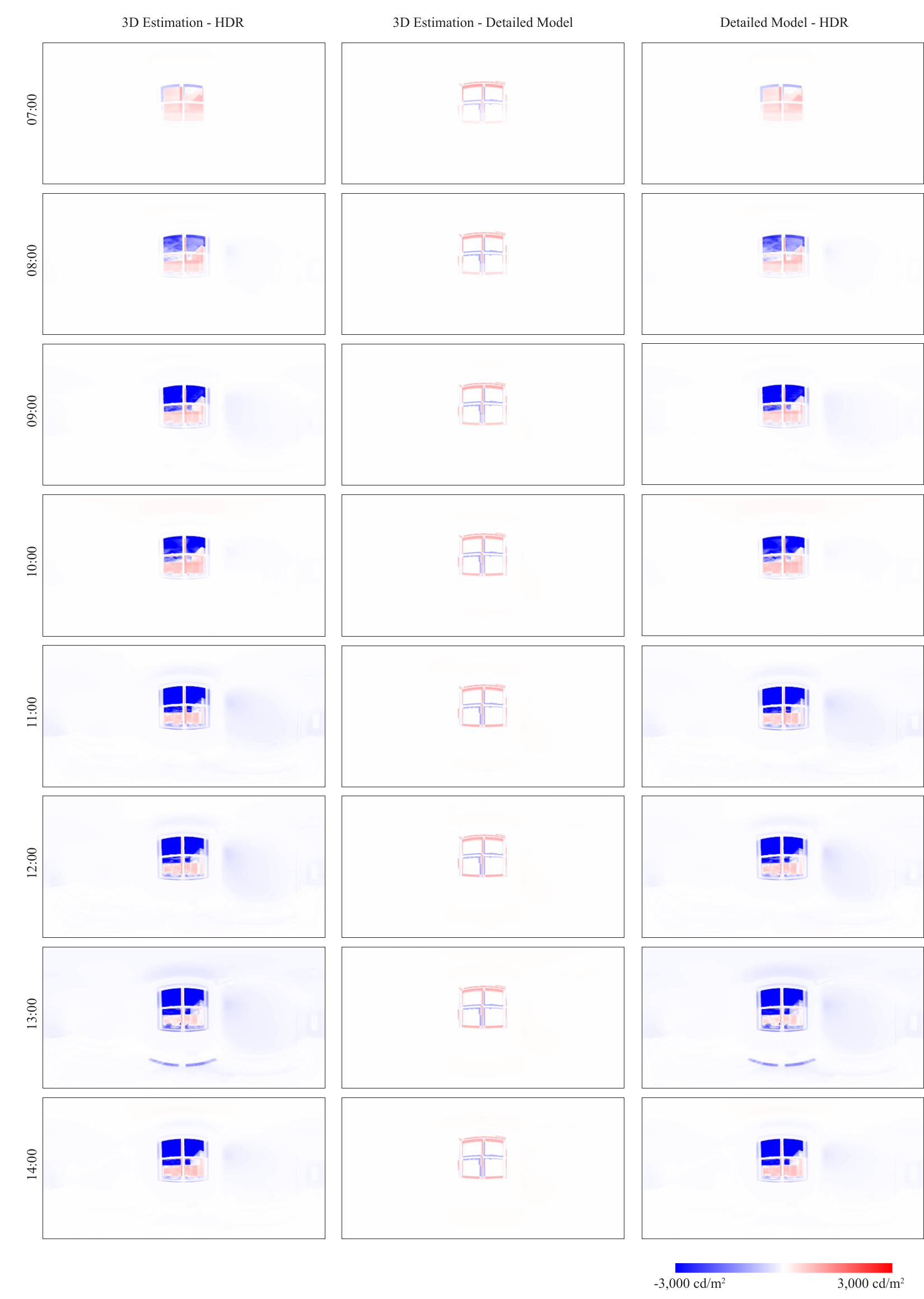}
    \caption{Panoramic luminance error maps comparing HDR photographs, 3D Estimation, and Detailed Models on June 22nd, 2023, under overcast sky conditions.}
    \label{fig:error_0622}
\end{figure*}

The error analysis was conducted on a per-pixel luminance comparison ranging from - 3,000 $cd/m^{2}$ to 3,000 $cd/m^{2}$, where blue indicates underestimation and red represents overestimation, respectively. We focus on clear sky condition on July 6th, 2023 (Figure.~\ref{fig:error_0706}) and overcast sky condition on June 22nd, 2023 (Figure.~\ref{fig:error_0622}).

In the clear sky scenario, after the error analysis of the 3D Estimation method and HDR Photography method (Figure.~\ref{fig:error_0706} (first column)), it was observed that the 3D estimation method caused higher luminance errors in outdoor scenes, and the ceiling consistently shows overestimation of luminance value, due to the absence of surrounding context in the simulation. For the wall and floor surfaces, minor errors occurred when no direct sunlight came from the outdoor scene. When direct sunlight appears on the indoor surfaces, the planar window surfaces in the 3D estimation model consistently bring extra sunlight region on the wall and floor surfaces. 

When comparing the 3D Estimation approach and the Detailed Model approach (Figure.~\ref{fig:error_0706} (second column)), the primary luminance differences are located around the window frame area. The 3D estimation method consistently brings larger direct sun regions (over 3,000 cd/$m^{2}$) into the indoor spaces. For other surfaces (like walls, floors, and ceilings), luminance error is small when there is no direct sunlight within the space.

When comparing the Detailed Model approach and HDR Photography approach (Figure.~\ref{fig:error_0706} (third column)), similar to digital estimation, the detailed model also exhibits errors in outdoor scenes due to a lack of surrounding context. The luminance value on the the ceiling is also overestimated in computational lighting simulation, due to the absence of outdoor context. The overestimation of the direct sun region on the wall and floor is less in the detailed model method, due to the presence of wall thickness and detailed window frame. 

Under overcast sky conditions (Figure.~\ref{fig:error_0622}), the luminance errors in the indoor space are smaller than in clear sky scenarios (Figure.~\ref{fig:error_0706}). Without outdoor context in the computational simulation, errors were mainly concentrated on the window region.

\subsection{Fisheye Luminance Map}
\begin{figure*}[!ht]
    \centering
    \includegraphics[width=0.70\linewidth]{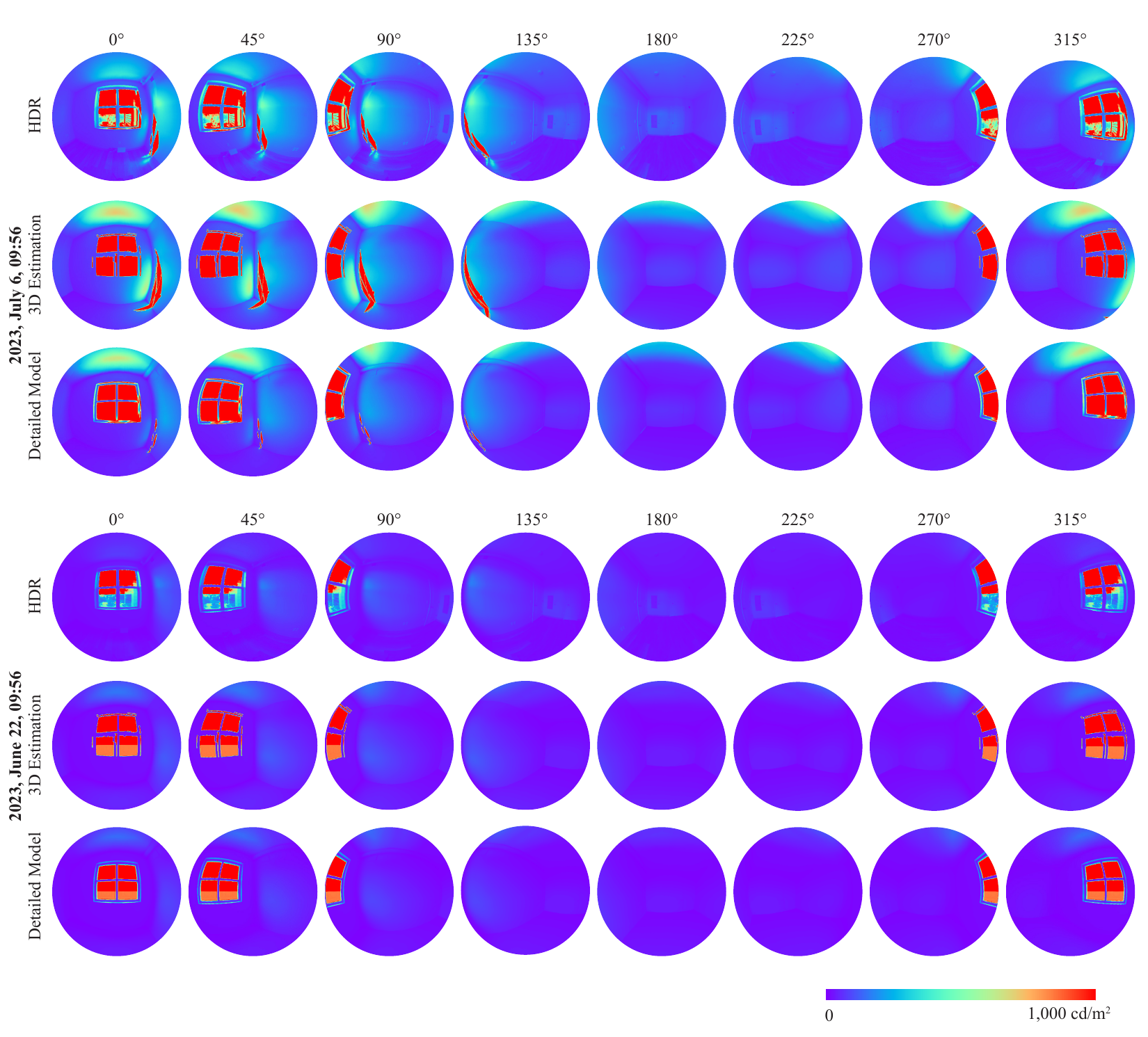}
    \caption{180$^{\circ}$ fisheye images in the false color scheme (from 0 to 1,000 cd/$m^{2}$) segmented from HDR photographs, 3D Estimation, and Detailed Models under clear sky conditions (July 6, 2023) and overcast sky conditions (June 22, 2023).}
    \label{fig:false_fisheye}
\end{figure*}

Except for comparing panoramic luminance maps from three methods, Figure.~\ref{fig:false_fisheye} presents the point-in-time results in fisheye perspective under clear sky condition on July 6th, 2023, at 09:56 and overcast sky condition on June 22nd, 2023, at 09:56. The results are displayed in false color ranging from 0 to 1,000 $cd/m^{2}$. For HDR photography, each panoramic image was converted to eight fisheye images with a 45$^{\circ}$ increment. The results of the 3D estimation method and the detailed model method are generated from computational simulation and displayed in 180$^{\circ}$ fisheye perspective side by side.  

In the clear sky scenario (July 6th, 2023, 09:56), compared to HDR photography and detailed model, the 3D estimation method shows larger areas with over 3,000 cd/$m^{2}$ luminance values. Meanwhile, when the window is within the field of view, the outdoor context, wall, and ceiling exhibit greater brightness due to the absence of outdoor structures blocking the incoming light in the 3D estimation model. When the window is outside the field of view, the fisheye images from the three methods show low luminance errors. 

In the overcast sky scenario (June 22nd, 2023, 09:56), the luminance differences between the three methods are smaller than in clear sky conditions. The 3D estimation method and the detailed model have close luminance values in the ceiling area.

\subsection{Glare Metric}
Except for the luminance errors in panoramic representation and fisheye perspective, we also compared the numerical difference using Daylight Glare Probability (DGP)~\cite{wienold2006evaluation} for glare simulation. DGP is calculated as:

{\fontsize{10}{12}\selectfont
\begin{equation}
    \mathrm{DGP}=5.87 \times 10^{-5} \mathrm{Ev}+9.8 \times 10^{-2} \log \left(1+\sum_{\mathrm{i}} \frac{\mathrm{L}_{\mathrm{s}, \mathrm{i}}^2 \omega_{\mathrm{s}, \mathrm{i}}}{\mathrm{Ev}^{1.87} \mathrm{P}_{\mathrm{i}}^2}\right)+0.16.
\end{equation}
}

where $L_{s,i}$ refers to the source luminance ($cd/m^2$), $\omega_{s,i}$ points to the solid angle, $E_v$ means the vertical illuminance (lux), and $P_i$ is the Guth position index for the field of view. 

Computational lighting simulation provides flexibility in calculating the potential glare problems on an annual basis. Given the annual glare simulation, we selected hourly glare results on March 21st, June 21st, September 21st, and December 21st. The target time interval on each day spans hourly from 08:30 to 14:30. As shown in Figure.~\ref{fig:room}, the view division lists from \#1 to \#16, which represents the rotating angle from 0$^{\circ}$ to 360$^{\circ}$ in 22.5$^{\circ}$ gap starting from the view \#1.

\begin{figure*}[!ht]
    \centering
    \includegraphics[width=0.9\linewidth]{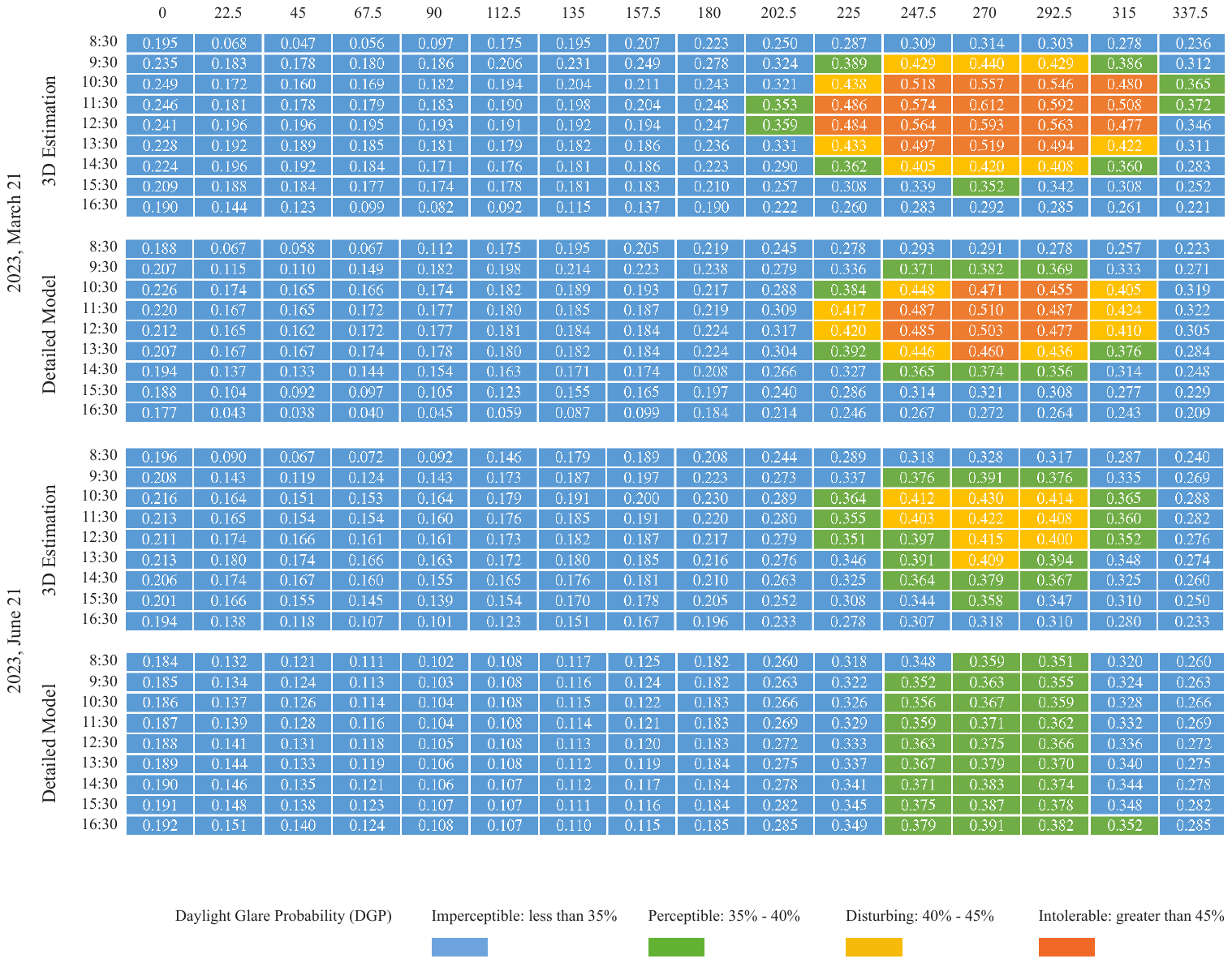}
    \caption{Results of glare metric using 3D Estimation and Detailed Model on March 21st, 2023, and June 21st, 2023, respectively.}
    \label{fig:enter-Glare1}
\end{figure*}

\begin{figure*}[!ht]
    \centering
    \includegraphics[width=0.9\linewidth]{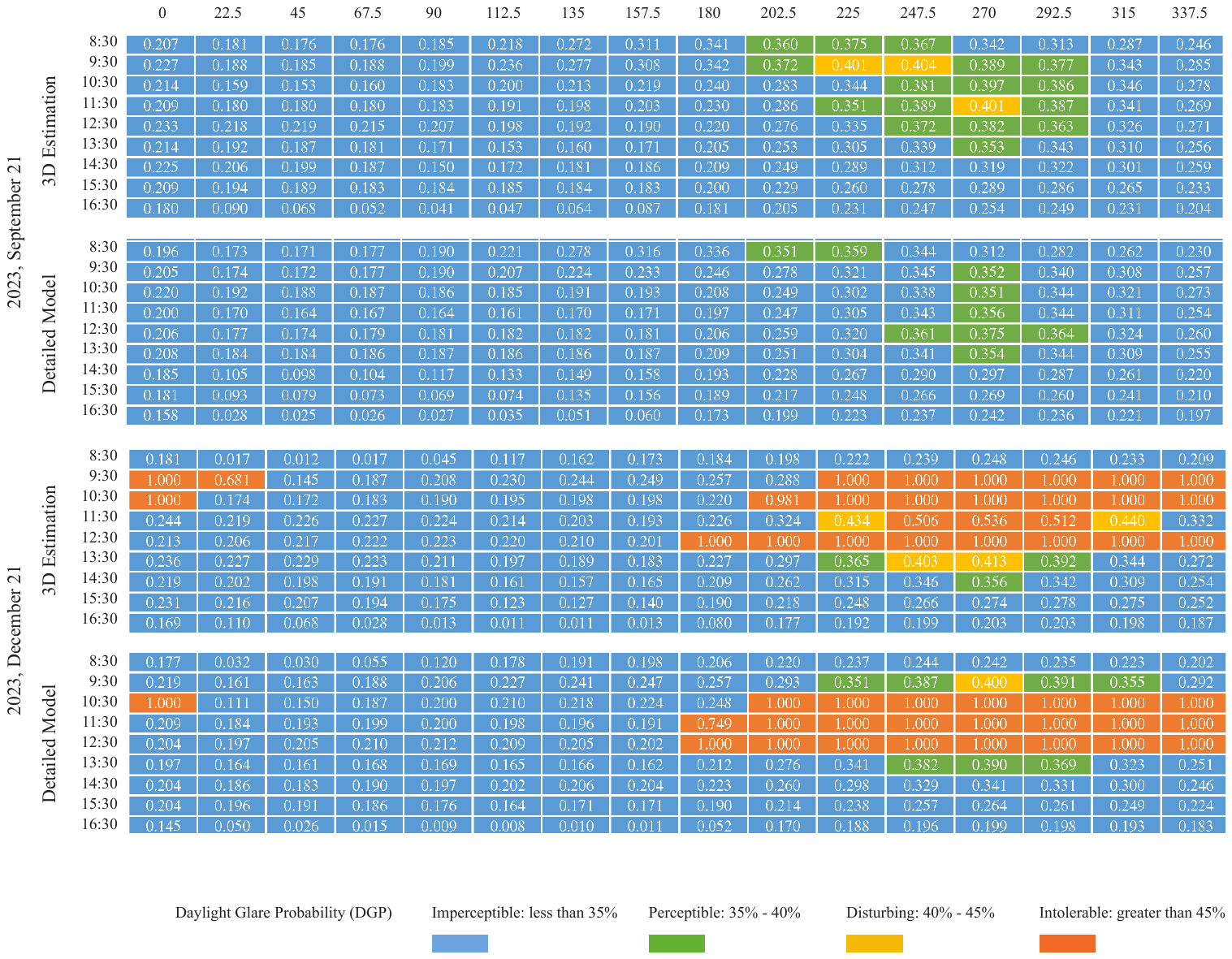}
    \caption{Results of glare metric using 3D Estimation and Detailed Model on September 21st, 2023, and December 21st, 2023, respectively.}
    \label{fig:enter-Glare2}
\end{figure*}

On March 21st (Figure.~\ref{fig:enter-Glare1} (top)), when the window is within the field of view, the DGP shows higher levels around noon time than other time points. Compared to the detailed model, the 3D estimation method results in higher glare levels due to the lack of detailed window structure and wall thickness. On June 21st (Figure.~\ref{fig:enter-Glare1} (bottom)) and September 21st (Figure.~\ref{fig:enter-Glare2} (top)), only perceptible glare is observed from the detailed model, while higher DGP values in Disturbing glare are generated from the 3D estimation model. On December 21st (Figure.~\ref{fig:enter-Glare2} (bottom)), the DGP value exhibits high values when the sun is visible from the window.

After comparing the DGP values from four typical dates, the 3D estimation method can identify the potential glare as the conventional detailed model approach, while the simplified wall and window surfaces consistently bring extra sunlight into the room and cause higher glare levels.

\section{Conclusion}
After evaluating the luminance maps from captured HDR photographs, the 3D Estimation model, and the Detailed Model, the results demonstrate that the 3D estimation method can accurately reconstruct room geometry. However, indoor scene details are simplified as planar surfaces, and detailed window frames and structures cannot be included in the 3D model. This research presents panoramic 3D estimation as an approach for generating room geometry to support lighting simulation. By comparing results in panoramic representation and fisheye perspective, it demonstrates that HDR photography, the 3D estimation method, and the detailed model have lower luminance errors on overcast days, whereas more errors occur when direct sunlight enters from the window under clear sky conditions.

\section{Limitation and Future Work}
This study has several limitations. First, it only focused on a single room to compare results with different weather conditions and times of the day. Future studies will include other indoor scenes with more complicated spaces, such as kitchens, living rooms, and other functional rooms. Second, only two dates and time-series panoramas were selected from the existing dataset. On-site HDR photography for other dates and time points will be necessary to further compare the luminance error. Third, the existing HDR images were collected in Pittsburgh, USA, and other geographic locations should also be considered.

\bibliographystyle{scsproc}

\bibliography{main}

\section*{Author Biographies}

\textbf{\uppercase{Zining Cheng}} is an Architecture Designer at Perkins\&Will, USA. Her research interests are modeling and simulation with applications in optimizing the indoor built environment for energy efficiency and human comfort. Her email address is \email{zining.cheng@perkinswill.com}.

\textbf{\uppercase{Guanzhou Ji}} is a PhD candidate at Carnegie Mellon University, USA. His research interests are indoor photometry, computational photography, and scene editing. His email address is \email{gji@andrew.cmu.edu}.

\end{document}